\documentclass[preprint]{iris-ai-25}
\usepackage[utf8]{inputenc}
\usepackage{orcidlink,thumbpdf,lmodern}
\usepackage{amssymb,amsmath,amsfonts,amsopn}
\usepackage{geometry}
\usepackage{framed}
\usepackage{xspace}
\usepackage{mdframed}
\usepackage{cleveref}
\usepackage[frozencache=true,cachedir=minted-cache]{minted}
\usepackage{microtype}
\usepackage{mathpazo}
\newcommand{\skfolio}{\texttt{skfolio}\xspace}
\newcommand{\sklearn}{\texttt{scikit-learn}\xspace}

\DeclareMathOperator*{\argmax}{argmax}
\DeclareMathOperator*{\argmin}{argmin}

\author{%
  Carlo Nicolini\corrauthor\equalcontribution \\
  Ipazia SpA \\
  Milan, Italy \\
  \texttt{c.nicolini@ipazia.com}\orcidlink{0000-0001-9661-0329} \\
  \And
  Matteo Manzi\equalcontribution \\
  Orion Finance\\
  Paris, France \\
  \texttt{matteomanzi09@gmail.com}\orcidlink{0000-0002-5229-0746} \\
  \And
  Hugo Delatte\equalcontribution \\
  SKFolio Labs \\
  London, UK \\
  \texttt{hugo.delatte@gmail.com} \\
}

\title{\skfolio: Portfolio Optimization in Python}
\date{\today}
\begin{document}

\maketitle

\begin{abstract}
Portfolio optimization is a fundamental challenge in quantitative finance, requiring robust computational tools that integrate statistical rigor with practical implementation. 
We present \skfolio, an open-source Python library for portfolio construction and risk management that seamlessly integrates with the scikit-learn ecosystem. 
\skfolio provides a unified framework for diverse allocation strategies, from classical mean-variance optimization to modern clustering-based methods, state-of-the-art financial estimators with native interfaces, and advanced cross-validation techniques tailored for financial time series.
By adhering to scikit-learn's fit-predict-transform paradigm, the library enables researchers and practitioners to leverage machine learning workflows for portfolio optimization, promoting reproducibility and transparency in quantitative finance.
\end{abstract}

\section{Introduction}

Portfolio optimization, first introduced by Markowitz's Modern Portfolio Theory (MPT) \citep{markowitz1952portfolio}, is a key concept in quantitative finance. However, putting MPT into practice comes with several challenges. 
These include high sensitivity to expected return and risk estimates, lack of diversification, frequent changes in asset weights (i.e. high turnover), and poor performance when tested on new data (i.e. overfitting). 
There are many different optimization methods, pre-selection techniques, distribution and moment estimators available, and they are often combined in practice, making the process even more complex.
This highlights the need for a single, unified framework that leverages machine learning for model selection, validation, and parameter tuning, while also mitigating the risks of overfitting and data leakage~\citep{arnott2019,bailey2016}. 

Such a framework is essential for the quantitative finance community, which faces persistent challenges in bridging the gap between sophisticated financial theory and practical implementation. 
These challenges include: (i) a fragmented ecosystem, where existing libraries lack consistent interfaces and comprehensive feature sets; (ii) limited reproducibility, as proprietary solutions hinder research validation and collaboration; (iii) poor machine learning integration, with traditional portfolio tools not fitting seamlessly into modern data science workflows; and (iv) inadequate cross-validation, since standard ML methods often fail to account for temporal dependencies inherent in financial data.

In this work, we present \href{https://github.com/skfolio/skfolio}{\skfolio}, an open-source library designed to address these challenges. Built on top of the standardized scikit-learn API~\citep{pedregosa2011scikit}, \skfolio\ offers a comprehensive framework for portfolio optimization and risk management. This approach ensures seamless integration with existing machine learning workflows, while preserving the mathematical rigor essential for robust portfolio construction.

\section{Design and Implementation}

\subsection{Overview and Basic Usage}

\begin{figure}[htb]
    \begin{mdframed}[linewidth=0.5pt]
    \begin{minipage}[t]{0.48\linewidth}
        {\small \sffamily A.}\\[1ex]
        \includegraphics[width=\linewidth]{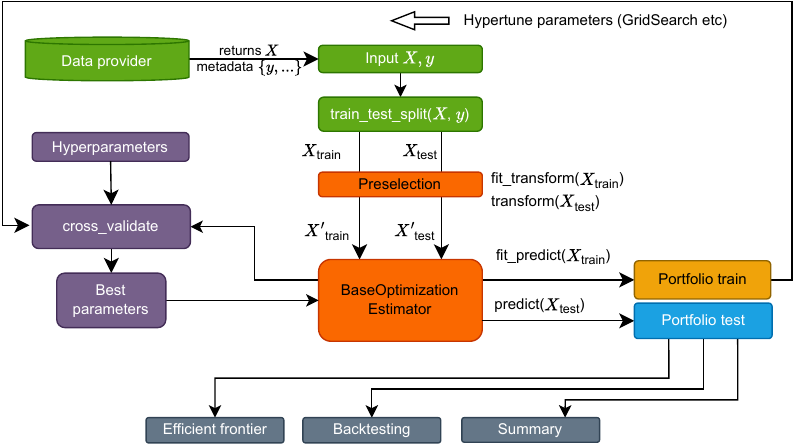}\\
        {\small \sffamily C.}\\[1ex]
        \includegraphics[width=0.9\linewidth]{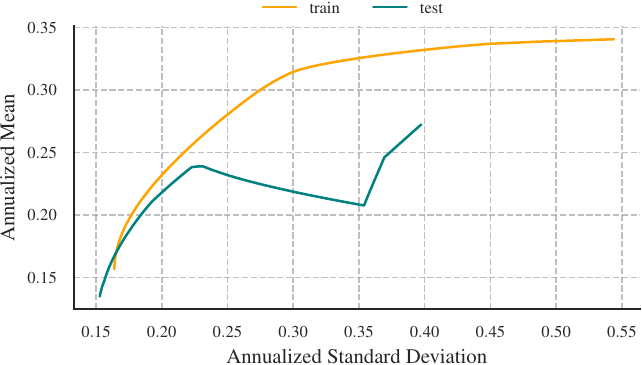}
    \end{minipage}%
    \hfill
    \begin{minipage}[t]{0.48\linewidth}
        {\small \sffamily B.}\\\vspace{-3ex}
        \begin{minted}[fontsize=\scriptsize]{python}
from skfolio.datasets import load_sp500_dataset
from skfolio.optimization import (
    MeanRisk, 
    ObjectiveFunction
)
from skfolio.measures import RiskMeasure
from skfolio.preprocessing import prices_to_returns
from sklearn.model_selection import train_test_split
# Load the prices and convert to returns
prices = load_sp500_dataset()
X = prices_to_returns(prices)
X_train, X_test = train_test_split(
    X, test_size=0.2, shuffle=False
)
# Define a convex optimization problem
model = MeanRisk(
  risk_measure=RiskMeasure.VARIANCE,
  objective_function=ObjectiveFunction.MINIMIZE_RISK,
  efficient_frontier_size=100,
  max_weights={"AAPL": 0.2},
  l2_coef=0.01
)

# Fit on training data and predict on test data
model.fit(X_train)
ptf_train = model.predict(X_train)
ptf_test = model.predict(X_test)
ptf_train.summary(), ptf_test.summary()
\end{minted}
    \end{minipage}
    \caption{(A) \skfolio architecture showing the integration between data preprocessing, optimization methods, and evaluation components within the scikit-learn framework. (B) Basic usage example of finding the minimum variance portfolio. (C) Efficient frontier for the minimum variance portfolio on the S\&P 500 dataset, over both train and test data. The training frontier dominates the test frontier, showing the overfitting of the model.}
    \label{fig:architecture}
    \end{mdframed}
\end{figure}

\skfolio follows a modular architecture that integrates seamlessly with the scikit-learn ecosystem. Figure~\ref{fig:architecture}A illustrates the library's core components and their interactions.
The library is built around three fundamental principles:
(i) \texttt{scikit-learn} compatibility: all estimators inherit from \href{https://scikit-learn.org/stable/modules/generated/sklearn.base.BaseEstimator.html}{\texttt{BaseEstimator}}, implementing the standard fit-predict-transform paradigm, (ii) mathematical rigor: state-of-the-art portfolio optimization methods with numerical stability implemented in cvxpy~\citep{diamond2016cvxpy}, (iii) transparency and reproducibility: open-source design promoting research reproducibility.

As an example, we demonstrate how to find the minimum variance portfolio with a maximum weight constraint and $L_2$ regularization on the portfolio weights, and how to generate the corresponding efficient frontier. 
This is formulated as a convex minimization problem.

\begin{equation}\label{eq:minimize_risk_example}
\begin{aligned}
\min_{\mathbf{w}} \quad & \mathbf{w}^T \boldsymbol{\hat{\Sigma}} \mathbf{w} + \lambda \mathbf{w}^T \mathbf{w} \\
\text{s.t.} \quad& \mathbf{w}^T \mathbf{1} = 1 \\
& \mathbf{w}_{\rm{AAPL}} \leq 0.2
\end{aligned}
\end{equation}
where $\boldsymbol{\hat{\Sigma}}$ is the sample covariance matrix, $\lambda$ is the $L_2$ regularization parameter.
In Figure~\ref{fig:architecture}B, we implement the convex optimization problem from~\Cref{eq:minimize_risk_example}.
We load the S\&P 500 dataset, convert prices to returns, and solve for 100 points on the mean-risk frontier, imposing a maximum weight of 20\% for \texttt{AAPL} and $L_2$ regularization with $\lambda=0.01$. 
Figure~\ref{fig:architecture}C, shows the frontier for the problem, divided over train and test set.

\subsubsection{Convex Optimization Methods}
The \texttt{MeanRisk} class covers all the convex optimization problems that can be formulated as risk minimization, expected returns maximization, utility maximization, and ratio maximization (Table~\ref{tab:convex_problems}). 
Multiple risk functions are supported (e.g. variance, CVaR, CDaR, and many more). 
All problems have an additional regularization cost described by the $\mathcal{L}(\mathbf{w})$ term including $L_1$ and $L_2$ norms of the portfolio weights, transaction costs, management fees and parameters uncertainty loss for robust parameters estimation~\citep{mohajerin2018data,ceria2006incorporating}.
Granular control over the number of assets in a portfolio can be achieved by applying cardinality constraints (i.e., specifying a maximum number of assets), which are supported through specialized MIP solvers. Prior estimators may also be incorporated to assist in estimating expected returns, the covariance matrix, or to impose additional structure on the optimization problem.

\begin{table*}[htb]
    \centering
    \begin{tabular}{p{0.48\textwidth}p{0.48\textwidth}}
        \textbf{Minimize Risk} &
        \textbf{Maximize Expected Returns}  \\
        $\displaystyle
        \begin{array}{l}
            \left\{
            \begin{array}{lcll}
            \argmin_{\mathbf{w}} & \rm{risk}_i(\mathbf{R}^T\mathbf{w}) + \mathcal{L}(\mathbf{w}) & & \\
            \text{s.t.} & \mathbf{w}^T{\boldsymbol \mu} \geq \mu_{\min} & & \\
            & \mathbf{A} \mathbf{w} \geq \mathbf{b} & & \\
            & \rm{risk}_{j}(\mathbf{R}^T\mathbf{w}) \leq \rm{risk}_j^{\max}, \forall \; i\neq j & &
            \end{array}
            \right.
        \end{array}
        $
        &
        $\displaystyle
        \begin{array}{l}
            \left\{
            \begin{array}{lcll}
            \argmax_{\mathbf{w}} & \mathbf{w}^T{\boldsymbol \mu} - \mathcal{L}(\mathbf{w}) & & \\
            \text{s.t.} & \mathbf{A} \mathbf{w} \geq \mathbf{b} & & \\
            & \rm{risk}_{j}(\mathbf{R}^T\mathbf{w}) \leq \rm{risk}_j^{\max},  \forall \; j & &
            \end{array}
            \right.
        \end{array}
        $ \\
        \\
        \textbf{Maximize Utility}  &
        \textbf{Maximize Ratio} \\
        $\displaystyle
        \begin{array}{l}
            \left\{
            \begin{array}{lcll}
            \argmax_{\mathbf{w}} & \mathbf{w}^T\mu - \lambda \times \rm{risk}_i(\mathbf{R}^T\mathbf{w}) - \mathcal{L}(\mathbf{w}) & & \\
            \text{s.t.} & \mathbf{w}^T\mu \ge \mu_{\min} & & \\
            & \mathbf{A} \mathbf{w} \geq \mathbf{b} & & \\
            & \rm{risk}_{j}(\mathbf{R}^T\mathbf{w}) \leq \rm{risk}_j^{\max},  \forall \; j & &
            \end{array}
            \right.
        \end{array}
        $ 
        &
        $\displaystyle
        \begin{array}{l}
            \left\{
            \begin{array}{lcll}
            \argmax_{\mathbf{w}} & \dfrac{\mathbf{w}^T{\boldsymbol \mu}}{\rm{risk}_{i}(\mathbf{R}^T\mathbf{w})} - \mathcal{L}(\mathbf{w}) & & \\
            \text{s.t.} & \mathbf{w}^T\mu \geq \mu_{\min} & & \\
            & \mathbf{A} \mathbf{w} \geq \mathbf{b} & & \\
            & \rm{risk}_{j}(\mathbf{R}^T\mathbf{w}) \leq \rm{risk}_j^{\max},  \forall \; j & &
            \end{array}
            \right.
        \end{array}
        $  \\
    \end{tabular}
    \caption{The four formulations of convex portfolio optimization problems implemented in \skfolio. The risk function $\rm{risk}_i$ is chosen from one of the many risk functions available. The optimization variables are the portfolio weights $\mathbf{w}$, the constraints are encoded in the $\mathbf{A}$ and $\mathbf{b}$ matrices. The main input variable are the asset excess returns $\mathbf{R}$. The parameter $\mu_{\min}$ is the minimum acceptable excess return, $\rm{risk}_i^{\max}$ is the maximum acceptable risk for the $i$-th risk measure, while the $\mathcal{L}(\mathbf{w})$ encodes additional regularization terms.}
    \label{tab:convex_problems}
\end{table*}

\subsection{Prior Estimation Methods}
Prior estimators are all the necessary methods to provide additional information to the optimization process.
\skfolio supports sophisticated prior estimation methods that enhance portfolio optimization by incorporating domain knowledge or analyst views, and addressing estimation errors.
Frequentist or Bayesian approaches like parameters shrinkage (Bayes-Stein for returns~\citep{jorion1986bayes}, shrinkage for covariance matrix~\citep{ledoit2004well}) and Black-Litterman~\citep{he2002intuition} are available as prior specifiers, together with classical factor models~\citep{jurczenko2015risk}. 
Advanced information-theoretic approaches, such as opinion pooling~\citep{good1952rational} and entropy pooling~\citep{meucci2011fully,meucci2012effective} are also supported, offering robust alternatives for asset allocation in non-Gaussian markets, where investors hold complex views beyond simple return forecasts.

The following example highlights how seamlessly \skfolio integrates with \sklearn by specifying a factor model prior within a convex optimization problem. Here, a ridge regressor with $\alpha=0.1$ is used to estimate the factor loading matrix, embedded directly in the \texttt{MeanRisk} optimizer. 
This setup allows for flexible model specification and, if desired, the ridge regularization parameter $\alpha$ can be further fine-tuned using time-series-aware cross-validation~(\Cref{sec:model_selection}).

\begin{minted}[fontsize=\scriptsize]{python}
from skfolio.prior import FactorModel, LoadingMatrixRegression
from sklearn.linear_model import Ridge
from skfolio.datasets import load_factors_dataset
factor_prices = load_factors_dataset()
# factors and prices are aligned in time
X, y = prices_to_returns(prices, factor_prices)
X_train, X_test, y_train, y_test = train_test_split(
    X, y, test_size=0.2, shuffle=False
)
# Create factor model with custom regression
model = MeanRisk(
    risk_measure=RiskMeasure.VARIANCE,
    objective_function=ObjectiveFunction.MINIMIZE_RISK,
    max_weights={"AAPL": 0.2},
    l2_coef=0.01,
    prior_estimator=FactorModel(
        loading_matrix_estimator=LoadingMatrixRegression(
            linear_regressor=Ridge(alpha=0.1),
        )
    )
)
model.fit(X_train, y_train)
\end{minted}

\subsubsection{Expected Returns and Covariance Estimation}
As the empirical covariance matrix often suffers from instability and estimation error, \skfolio addresses this by implementing a comprehensive suite of advanced covariance estimators.
Shrinkage methods improve stability by contracting the empirical matrix toward a structured target~\citep{ledoit2004well}.
Techniques based on Random Matrix Theory (RMT)~\citep{laloux2000random} denoise the matrix by filtering out eigenvalues associated with noise.
For uncovering sparse dependency structures, the Graphical Lasso~\citep{friedman2008sparse} is available for estimating the inverse covariance matrix.
The library also includes other robust estimators, like the Gerber statistic~\citep{gerber2019gerber}, which measures co-movement by focusing only on significant joint fluctuations.

Estimation of expected returns implementing shrinkage~\citep{jorion1986bayes} and exponentially weighted averaging can also be used to improve the stability of solutions.

\subsubsection{Copula-based Synthetic Priors}

\begin{figure}[htb]
    \centering
    \begin{mdframed}[linewidth=0.5pt]
    \includegraphics[width=0.48\linewidth]{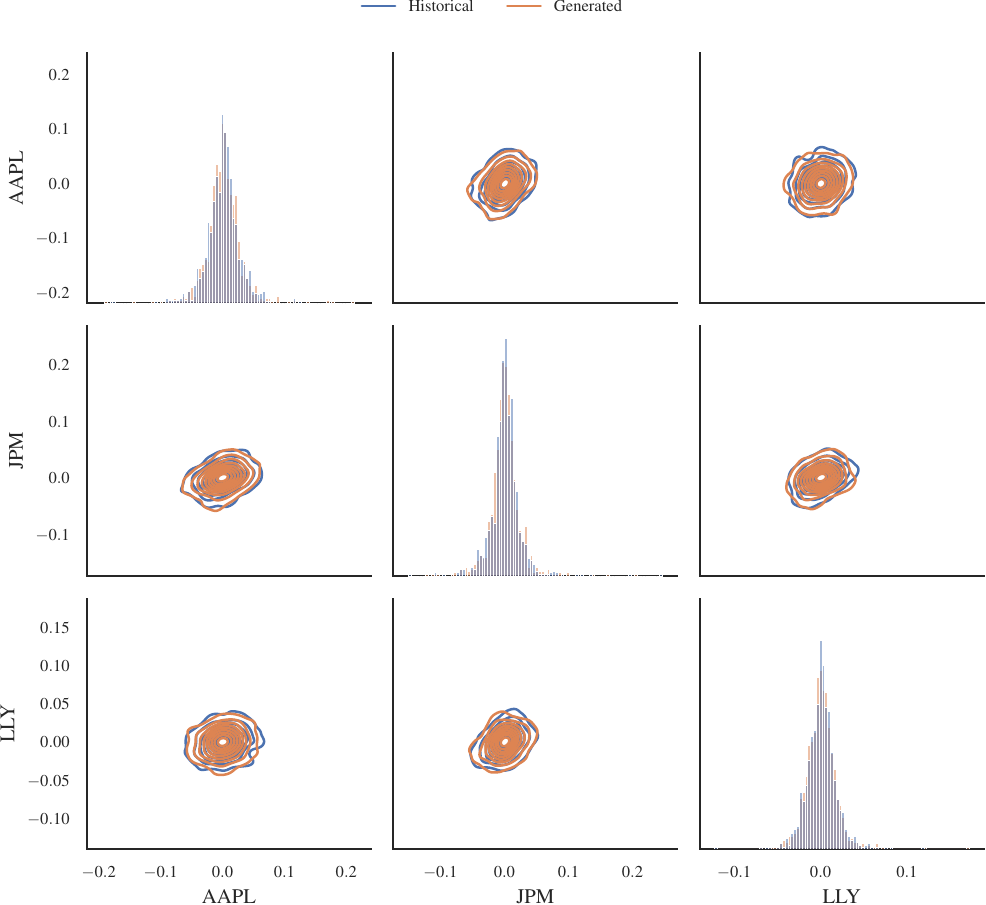}
    \includegraphics[width=0.48\linewidth]{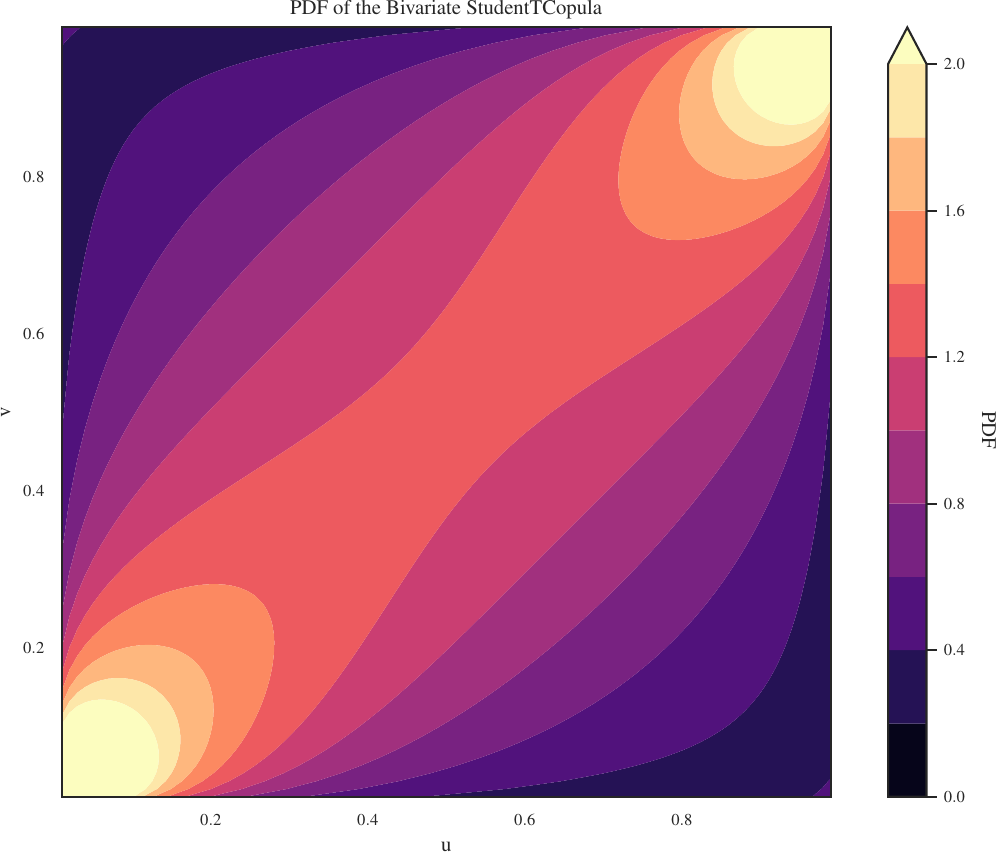}
    \end{mdframed}
    \caption{Left panel: joint distribution visualization of the vine copula structure and synthetic data generation results over just three asset pairs. The synthetic data are generated by sampling from a vine copula fitted on the historical returns. Right panel: two dimensional density of a bivariate Student-t copula on historical \texttt{AAPL} and \texttt{JPM} returns.}
    \label{fig:vine_copula_plots}
\end{figure}

\skfolio provides advanced stress testing capabilities through synthetic data generation using copula models. 
In recent years, copula-based methods~\citep{nelsen2006introduction} have emerged as a powerful alternative to traditional mean-variance approaches in portfolio optimization~\citep{mcneil2015quantitative}, particularly when modeling nonlinear dependencies, tail risk, and asymmetric correlations~\citep{kakouris2014robust,han2017dynamic}. 
Common bivariate copula models are supported, (Gaussian, Student-t, etc.), together with the multi-variate vine copula, crucial for capturing real-world phenomena such as heavy tails, skewness, and extreme co-movements—features frequently observed in financial time series but poorly captured by multivariate normal distributions.
Additional conditioning on vine copula allows to impose views on the tail dependencies of the returns distribution, as shown in Figure~\ref{fig:vine_copula_plots}.

\subsection{Ensemble and clustering}
\skfolio includes alternative optimization approaches that do not rely solely on convex optimization. Many of these methods leverage clustering or ensemble techniques. 
For example, Hierarchical Risk Parity~\citep{lopez2016building} builds portfolios without requiring inversion of the covariance matrix, improving stability. Nested Clustering Optimization~\citep{lopez2016robust} further enhances robustness by combining hierarchical clustering with cross-validation to reduce overfitting.
In the context of financial markets, where data is often characterized by noise and non-stationarity, ensemble techniques like stacking~\citep{wolpert1992stacked,de2018advances} are also very effective in mitigating overfitting and enhancing portfolio robustness and out of sample performance.

\section{Model selection}\label{sec:model_selection}

Financial data often exhibit serial correlation, meaning that past values are correlated with future values, and this characteristic can lead to overfitting if not properly addressed.
Moreover, financial datasets are prone to train-test leakage, where information from the test set inadvertently influences the training process, thus skewing the evaluation of the model's performance.

To mitigate these issues, \skfolio employs advanced cross-validation techniques such as Combinatorial Purged Cross-Validation (CPCV) with purging and embargoing~\citep{de2018advances}. 
Unlike traditional KFold cross-validation, which randomly splits the data into folds, CPCV is designed to handle the temporal dependencies inherent in time series data. 
It does so by using $k - p$ folds for training, where $p > 1$ allows for multiple test folds, thereby providing a more robust evaluation of the model's predictive power.
Purging involves removing any training data that overlaps in time with the test set, ensuring that no future information is leaked into the training process.
Embargoing further enhances this by excluding data immediately following the test set from the training data, preventing any potential leakage from adjacent time periods.

Additionally, a walk-forward cross-validator implements a rigorous, time-ordered splitting strategy tailored for portfolio backtesting.
In contrast to conventional k-fold schemes, it respects the chronological order of observations, ensuring that each test set consists solely of future data relative to its training counterpart. 
This design provides a high degree of control for evaluating portfolio strategies under realistic, forward-looking conditions.

In the following example we demonstrate the use of stacking optimization with three estimators based on convex and hierarchical methods.
We use a walk-forward cross-validation to evaluate the portfolio performance on the test set with the custom \texttt{cross\_val\_predict} function from \skfolio.

\begin{minted}[fontsize=\scriptsize]{python}
from skfolio import Population
from skfolio.optimization import (
  StackingOptimization,
  EqualWeighted, 
  HierarchicalRiskParity,
  InverseVolatility
)
from skfolio.model_selection import WalkForward, cross_val_predict
estimators = [
  ("IV", InverseVolatility()),
  ("CVAR", MeanRisk(risk_measure=RiskMeasure.CVAR)),
  ("HRP", HierarchicalRiskParity()),
]
benchmark = EqualWeighted().fit(X_train)
model_stacking = StackingOptimization(
  estimators=estimators,
  final_estimator=MeanRisk(
      risk_measure=RiskMeasure.CVAR,
  )
)
cv = WalkForward(train_size=252, test_size=60)
pred_stacking = cross_val_predict(
  model_stacking, X_test, cv=cv, 
  portfolio_params={"name": "Stacking"}
)
pred_benchmark = cross_val_predict(
  benchmark, X_test, cv=cv,
  portfolio_params={"name": "EW"}
)
Population([pred_stacking, pred_benchmark]).summary()
\end{minted}

\section{Software Architecture and Availability}

\skfolio is implemented in Python 3.10+ and distributed under the BSD 3-clause license. 
Key dependencies include NumPy, SciPy, scikit-learn, cvxpy-base~\citep{diamond2016cvxpy} for convex optimization and Clarabel~\citep{goulart2024clarabel} for interior-point solvers.
Other commercial solvers like Gurobi~\citep{gurobi}, MOSEK~\citep{mosek} and CPLEX~\citep{cplex2009v12} can be used as well.
The library is available via PyPI and conda-forge, with comprehensive documentation at \url{https://skfolio.org}.
The codebase follows established software engineering practices: comprehensive test suite with $>95\%$ code coverage, continuous integration via GitHub Actions, automated documentation generation, type hints and docstring standards for maintainability.

\section{Ongoing Innovation and Development}

The \skfolio library is under continuous, active enhancement, with a focus on integrating the latest advances in portfolio theory. We are currently implementing the Schur Complementary Allocation method \citep{cotton2024schur} and adding support for Multiple Randomized Cross-Validation \citep{palomar2025}. These ongoing efforts ensure that \skfolio remains a living, rapidly evolving toolkit for both researchers and practitioners.

\section{Conclusion}

\skfolio bridges the gap between sophisticated portfolio optimization theory and practical implementation by providing a comprehensive, open-source framework integrated with the scikit-learn ecosystem. Its unified API, advanced cross-validation methods, and emphasis on reproducibility make it a valuable tool for researchers and practitioners in quantitative finance.

The library's modular design and extensive documentation lower barriers to entry while maintaining mathematical rigor. By promoting transparency and reproducibility, \skfolio contributes to the democratization of advanced portfolio optimization techniques.

Future development focuses on expanding the library's capabilities while maintaining its core design principles of simplicity, transparency, and integration with the broader Python scientific computing ecosystem.

\begin{ack}
We thank the \skfolio open-source community and researchers for their contributions and feedback.  
Special recognition goes to the scikit-learn and CVXPY contributors for their direct work on the tools that underpin this library, as well as to the authors and maintainers of Riskfolio-Lib \citep{riskfolio} and PyPortfolioOpt \citep{Martin2021} for the inspiration they provided.  
We also especially thank Vincent Maladière for his detailed feedback and suggestions.

\end{ack}

\small
\bibliographystyle{plainnat}
\bibliography{biblio}

\end{document}